\documentclass{article}
\usepackage{spconf,amsmath,epsfig,subcaption,hyperref,url}

\title{Regression Networks For Calculating Englacial Layer Thickness}
\name{Debvrat Varshney\textsuperscript{1}, Maryam Rahnemoonfar\textsuperscript{1}, Masoud Yari\textsuperscript{1}, John Paden\textsuperscript{2} \thanks{This work is supported by NSF BIGDATA awards (IIS-1838230, IIS-1838024), IBM, and Amazon.}}
\address{\textsuperscript{1}Computer Vision and Remote Sensing Laboratory, University of Maryland Baltimore County, MD, USA\\\textsuperscript{2}Center of Remote Sensing of Ice Sheets (CReSIS), University of Kansas, KS, USA}
\begin{document}
%\ninept
%
\maketitle
\begin{abstract}
Ice thickness estimation is an important aspect of ice sheet modelling. In this work, we use convolutional neural networks (CNN) with multiple output nodes to regress and learn the thickness of internal ice layers in Snow Radar images captured over northwest Greenland. We experiment with some state-of-the-art CNNs to obtain a mean absolute error of 1.251 pixels of thickness estimation over the test set. Such regression-based networks can further be improved by embedding domain knowledge and radar information in the neural network in order to reduce the requirement of manual annotations.
\end{abstract}
\section{Introduction}
\label{sec:intro}
Rapidly changing climate is adversely affecting the world resulting in a negative ice sheet mass balance, a primary contributor to sea level rise. The Intergovernmental Panel on Climate Change (IPCC) reports \cite{stocker2014summary} that the sea level increase over the next century could potentially cause floods that risk millions of livelihoods. It is imperative to estimate the seasonal changes in ice sheets accurately, and build systems that would help in mitigating the possible calamity.

Annual snow accumulation rates can be calculated from detecting the englacial ice layer thickness \cite{https://doi.org/10.1002/grl.50706}. The internal ice layers can be detected with ground penetrating radar sensors, such as the Snow Radar flown for the NASA Operation IceBridge (OIB) mission. This sensor captures the vertical profile of an ice sheet where the varying contrast in radar reflections can be annually dated \cite{https://doi.org/10.1002/grl.50706}. The captured images showcase the most recent layer, the ice surface,  at the top and with layers from preceding years appearing underneath it. These images are noisy, and the multiple ice layers cannot be easily extracted even with modern image processing techniques \cite{koenig-accumulation-rate,macgregor2015}.

Convolutional Neural Networks (CNNs) are intelligent algorithms which have recently shown advantages for computer vision tasks such as image classification, object detection and semantic segmentation. In this paper, we use them for numerical regression to predict the thickness of each internal ice layer present in a Snow Radar image. This work is an improvement over \cite{varshney2020deep} where ice thickness was calculated as a separate, post-processing step after semantic segmentation of snow radar images. The rest of the paper is divided as follows. Section II describes the related work in ice layer detection with ground penetrating radar images and Section III describes the CNNs we use for regression. Section IV showcases the results and Section V concludes the paper.

\section{Related Work}
\label{sec:backgrnd}
%%%%%%% commented to save space
% In this Section, we discuss some of the past work on ice layer detection, and deep learning based regression tasks on images.

% \subsection{Ice Layer Detection} %%%% commented to save space

%%%%% ice layer and bedrock tracking
% In the past, multiple techniques have been proposed for detecting ice layers, such as, \cite{bruzzone-carrer,bruzzone1,level-set,charged-particles}. The authors of \cite{bruzzone-carrer} developed a hidden markov model to process planetary radargrams, while the authors of \cite{bruzzone1} coupled denoising methods with Steger and Weiner filtering to detect linear features of ice from radar data. Authors of \cite{level-set} evaluated airborne radar images with a level set approach, and the authors of \cite{charged-particles} used contour detection with anisotropic diffusion to identify ice and bedrock layers. These automated techniques detected only the ice surface and bedrock. 

In the past, techniques such as \cite{macgregor2015,koenig-accumulation-rate,dePaulOnana} were proposed to detect all the internal ice sheet layers. These were semi-supervised methods which required human correction, and their outputs were sparse in nature \cite{varshney2020deep}. Recent efforts such as \cite{yari-hed, yari2020multi-scale, rahnemoonfar2020radar, varshney2020deep} applied multi-scale deep learning techniques to track and identify internal ice layers in a pixel-wise manner. The purpose of tracking all internal layers is ultimately to calculate changes in accumulation rate over time. In our work, rather than creating image level pixel outputs, we build regression networks to directly learn and predict the thickness of ice layers, given an input radar image. By using well trained generalized networks, these models can be scaled to larger datasets captured from a variety of regions. The methodology of how we build these regression based CNNs is described in Section \ref{sec:method}.

% \subsection{Deep Regression of Images} %%%%% commented to save space
Some popular works on image regression have been \cite{rothe2018deep,doi:10.1080/21681163.2016.1149104} where the former used a VGG-16 architecture to predict the real and apparent age of a person from their facial image; whereas the latter created cell density maps through fully convolutional regression networks. For ice thickness estimation, works such as \cite{Shan:20,tc-13-2915-2019,rs8090698,kim2020prediction} were proposed. \cite{Shan:20} used a deep residual network to regress Raman spectral data, the output of which is later used for sea ice thickness measurements. \cite{tc-13-2915-2019} used a simple 3 layer CNN with a single output node to regress Antarctic sea ice thickness from lidar data, whereas \cite{rs8090698} used decision trees on CyroSat-2 and MODIS images to detect sea ice freeboard; which is useful in calculating sea ice thickness. \cite{kim2020prediction} also built a deep CNN with a single output node to calculate sea ice concentration, and compared its performance with random forest and a linear regressor.

In this work, we set up multi-output regression networks where each output node can estimate the thickness of each individual internal ice layer.

\section{Dataset}
\label{sec:dataset}
\begin{figure}[h!]
\centering
\begin{subfigure}{0.4\columnwidth}
  \centering
  \includegraphics[width=0.9\linewidth]{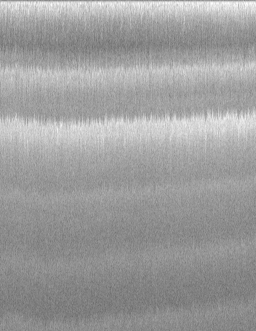}
  \caption{}
  \label{fig:sample-radar}
\end{subfigure}%
\begin{subfigure}{0.4\columnwidth}
  \centering
  \includegraphics[width=0.9\linewidth]{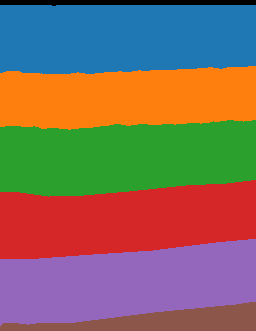}
  \caption{}
  \label{fig:sample-gt}
\end{subfigure}%
\begin{subfigure}{0.2\columnwidth}
  \centering
  \includegraphics[width=0.9\linewidth]{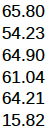}
  \caption{}
  \label{fig:sample-thick}
\end{subfigure}
\caption{(a) A sample snow radar image, (b) its labels from \cite{varshney2020deep} and (c) mean thickness of the labelled layers, in terms of number of pixels. %Each image is trained with 27 mean thickness values. In case an image does not have a particular layer, its thickness value is kept as 0.
}
\label{fig:sample-images}
\end{figure}

We use ultra-wideband Snow Radar data captured over northwest Greenland by the Center for Remote Sensing of Ice Sheets (CReSIS) in 2012 during the OIB Arctic campaign. This instrument has a vertical resolution of ~4cm per pixel in snow and is designed to detect shallow annual layers in the ice sheet. \cite{koenig-accumulation-rate}.
% \cite{cresis-ku}
This dataset contains 2361 training images along with 260 test images. Ice layers in this dataset were sparsely detected by \cite{koenig-accumulation-rate}, which were further processed by \cite{varshney2020deep} to remove incomplete layers. We use the latter work's processed layers for calculating the mean thickness of each layer, and treating these thickness values as our `ground truth' for regression. A sample Snow Radar image, its layers from \cite{varshney2020deep} and their mean thickness values are shown in Figure \ref{fig:sample-images}. This dataset contains a maximum of 27 detected layers, which has been pre-computed from the labelled dataset.

%%%%% sample image + semantic labels %%%%%%%%%
% \begin{figure}[h!]
% \centering
% \begin{subfigure}{0.5\columnwidth}
%   \centering
%   \includegraphics[width=0.9\linewidth]{figures/sampleimage_20120423_01_068__crop1.png}
%   \caption{}
%   \label{fig:sample-radar}
% \end{subfigure}%
% \begin{subfigure}{0.5\columnwidth}
%   \centering
%   \includegraphics[width=0.9\linewidth]{figures/samplelayer_20120423_01_068__crop1.png}
%   \caption{}
%   \label{fig:sample-gt}
% \end{subfigure}%
% \caption{Sample snow radar image (a), and its ground truth (b) prepared through \cite{varshney2020deep}}
% \label{fig:sample-images}
% \end{figure}

%%%%% with minipage package from IGARSS template %%%%%%
% \begin{figure}[htb]
% \begin{minipage}[b]{.48\linewidth}
%   \centering
% \centerline{\epsfig{figure=figures/sampleimage_20120423_01_068__crop1.png,width=4.0cm}}
% %   \vspace{1.5cm}
%   \centerline{(b) Results 3}\medskip
% \end{minipage}
% \hfill
% \begin{minipage}[b]{0.48\linewidth}
%   \centering
% \centerline{\epsfig{figure=figures/samplelayer_20120423_01_068__crop1.png,width=4.0cm}}
% %   \vspace{1.5cm}
%   \centerline{(c) Result 4}\medskip
% \end{minipage}
% %
% \begin{minipage}[b]{1.0\linewidth}
%   \centering
% \centerline{\epsfig{figure=figures/samplethickness_20120423_01_068__crop1.png,width=3.0cm}}
% %   \vspace{2.0cm}
%   \centerline{(a) Result 1}\medskip
% \end{minipage}
% \caption{Example of placing a figure with experimental results.}
% \label{fig:res}
% %
% \end{figure}

\section{Methodology}
\label{sec:method}
Regression networks are those that predict continuous values of a variable, rather than discrete labels as those predicted in classification networks. The predicted variables can take any value, including decimal values. For preparing our ground truth for regression, we calculated the thickness of each layer in every image from the ground truth (Figure \ref{fig:sample-gt}). For calculating the thickness of each layer, we first calculated the number of pixels for each label in the ground truth image, and then divided it by the total number of columns (width) of the image. This would give the mean number of rows per column for every label, which corresponds to the average thickness of each layer (Figure \ref{fig:sample-thick}). We also pre-calculated the maximum number of layers in our ground truth, which was found to be 27. 

\begin{figure}[h!]
    \centering
    \includegraphics[width=\columnwidth]{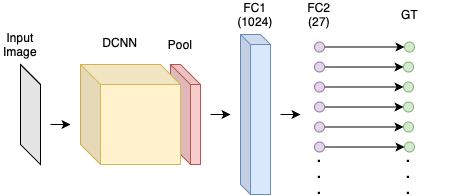}
    \caption{Regression network for calculating the thicknesses.}
    
    % CNN architecture that was used for regression. Here, baseline network is each of: InceptionV3 \cite{inceptionv3}, DenseNet \cite{densenet}, ResNet50 \cite{resnet50}, Xception \cite{xception} and MobileNetV2 \cite{sandler2018mobilenetv2}
    % }
    \label{fig:reg_net}
\end{figure}

Once we had the ground truth thickness values prepared, we built our regression network. We used classification networks like InceptionV3 \cite{inceptionv3}, DenseNet \cite{densenet}, ResNet50 \cite{resnet50}, Xception \cite{xception} and MobileNetV2 \cite{sandler2018mobilenetv2} as baseline models and added a global average pooling layer along with a fully connected layer of 1024 neurons, to the output of the baseline models. Finally, we added a fully connected layer of 27 nodes to each model, which corresponds to the number of layers and thickness values we want to predict in every image during inference. The overall architecture is shown in Figure \ref{fig:reg_net}. Here, `DCNN' is a deep convolutional neural network, such as InceptionV3, DenseNet, Xception etc. `FC' represents fully connected layers, with the number of nodes shown in the parenthesis. FC1 contains 1024 nodes and FC2 contains 27 nodes. `GT' is the ground truth in terms of thickness estimates. As this was a regression problem, instead of classification, we used a ReLU activation instead of Softmax. We also compared our work with the output of \cite{varshney2020deep} where thickness was calculated as a post-processing step.

% The rest of the section gives a brief description of each of the baseline networks and also describes our experimental setup.

\subsection{Baseline CNN models}

Each of the networks InceptionV3, DenseNet, ResNet50 \cite{resnet50}, Xception, and MobileNetV2  have significantly contributed to the field of computer vision by introducing a unique design element, such as the skip connections and residual blocks in ResNet, the multi-scale architecture in Inception and depthwise separable convolutions in Xception. DenseNet contains shorter connections between layers close to the input, and those layers close to the output. For each layer, feature maps of all preceding layers are used as input and thus help in feature propagation and reducing the vanishing gradient problem. MobileNetV2 helps in taking a low-dimensional compressed representation of the input, and filters it with lightweight depthwise separable convolutions. Such an architecture made it efficient for processing images through mobile devices.

\subsection{Experimental Setup}
All networks were trained for 100 epochs with the Adam optimizer \cite{DBLP:journals/corr/KingmaB14} having a learning rate of 0.0001 and a batch size of 32. They were trained to minimize the mean absolute error (MAE) given by Equation \ref{eq:mae}, where $p_i$ is the predicted thickness (in pixels) by the neural networks and $t_i$ is their thickness (in pixels) from ground truth labels. $k$ is the number of labels, which is 27 in our case. We used an NVIDIA GeForce RTX 2080 Ti GPU with an Intel Core i9 processor for our experiments.

\begin{equation}
MAE = \frac{\sum\limits_{i=1}^k\mid p_i-t_i \mid}{k}
\label{eq:mae}
\end{equation}

\section{Results and Disucssion}
\label{sec:results}
\begin{figure}[h!]
    \centering
    \includegraphics[width=\columnwidth]{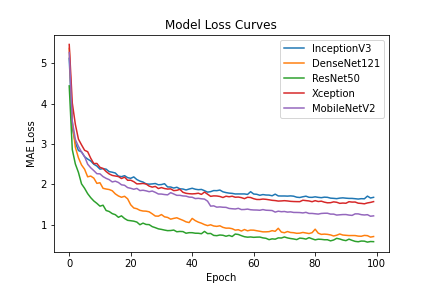}
    \caption{The networks marked in the legend were used as baseline models (DCNN) of Figure \ref{fig:reg_net}. The plot here shows their loss curves.}
    \label{fig:loss_curves}
\end{figure}
\begin{table}[h!]
    \centering
    \begin{tabular}{c c c}
         Baseline Model (DCNN) & Train & Test   \\
         \hline
         InceptionV3 \cite{inceptionv3} & 1.645 & 2.145\\
         DenseNet121 \cite{densenet} & 0.642 & 1.307\\
         ResNet50 \cite{resnet50} & \textbf{0.595} & \textbf{1.251} \\
         Xception \cite{xception} & 1.472 & 1.966\\
         MobileNetV2 \cite{sandler2018mobilenetv2} & 1.490 & 2.132\\
         post-DeepLabv3+\cite{varshney2020deep} & 2.36 & 3.59
    \end{tabular}
    \caption{Mean absolute error (MAE) in pixels computed over the training and test set through our regression networks.}
    \label{tab:mae_quant}
\end{table}

This section highlights the results that we achieved and gives a brief analysis. The loss curves of the five networks we trained are shown in Figure \ref{fig:loss_curves} and the MAE achieved by each of them is shown in Table \ref{tab:mae_quant}. This table also compares the MAE achieved by \cite{varshney2020deep} as the post-DeepLabv3+ model. From Figure \ref{fig:loss_curves} we see that ResNet50 trained much more quickly, achieving a lesser MAE before any other network. On the other hand, Table \ref{tab:mae_quant} shows that InceptionV3 gave the largest MAE followed by MobileNetV2, Xception, DenseNet121, and then ResNet50. InceptionV3 was built to cater to images having the same object of different sizes. Hence, the architecture was introduced with multiple 5$\times$5 and 3$\times$3 convolutions on the same input image. We note that although such a feature might be helpful for a dataset such as ImageNet \cite{5206848}, this is not useful for the CReSIS dataset where ice layers are generally of the same size and width. We see that ResNet50, with its skip connections in the residual blocks and DenseNet, with its densely connected network give the lowest (best) MAE values. Further, depthwise convolutions, present in Xception and MobileNetV2 architectures, were found to be much more useful than the inception modules of InceptionV3. Thus, residual learning, or densely connected networks, where each layer is connected with all of its succeeding layer helps in network learning and reducing the number of weights required. Such network strategies help in feature propagation and efficient extraction of image features.

Overall, obtaining a mean absolute error of approximately 0.6 to 1.2 pixels is good since it translates to just 2 to 5 centimeters of error in ice layer thickness, which are on the order of 1 meter in thickness. Thus, using regression networks has been especially useful, since they directly learn the layer thickness values, rather than first learning the spatio-contextual (pixel-wise) distribution of ice layers as in \cite{varshney2020deep}.

\section{Conclusion}
\label{sec:conc}
This work saw the successful use of CNNs in regressing ice layer thickness values, which is an important requirement for climate studies. Further, the skip or dense connections from ResNet and DenseNet were found to be more useful than depthwise convolutions or the inception module. Directly regressing for layer thickness values is much more useful than performing semantic segmentation and calculating thickness as a post-processing step. For future work, such regression networks can be combined with domain knowledge in order to retrieve accurate information with lesser number of labels.

% % Below is an example of how to insert images. Delete the ``\vspace'' line,
% % uncomment the preceding line ``\centerline...'' and replace ``imageX.ps''
% % with a suitable PostScript file name.
% % -------------------------------------------------------------------------
% \begin{figure}[htb]

% \begin{minipage}[b]{1.0\linewidth}
%   \centering
% % \centerline{\epsfig{figure=image1.ps,width=8.5cm}}
%   \vspace{2.0cm}
%   \centerline{(a) Result 1}\medskip
% \end{minipage}
% %
% \begin{minipage}[b]{.48\linewidth}
%   \centering
% % \centerline{\epsfig{figure=image3.ps,width=4.0cm}}
%   \vspace{1.5cm}
%   \centerline{(b) Results 3}\medskip
% \end{minipage}
% \hfill
% \begin{minipage}[b]{0.48\linewidth}
%   \centering
% % \centerline{\epsfig{figure=image4.ps,width=4.0cm}}
%   \vspace{1.5cm}
%   \centerline{(c) Result 4}\medskip
% \end{minipage}
% %
% \caption{Example of placing a figure with experimental results.}
% \label{fig:res}
% %
% \end{figure}

% % To start a new column (but not a new page) and help balance the last-page
% % column length use \vfill\pagebreak.
% % -------------------------------------------------------------------------
% \vfill
% \pagebreak

% References should be produced using the bibtex program from suitable
% BiBTeX files (here: strings, refs, manuals). The IEEEbib.bst bibliography
% style file from IEEE produces unsorted bibliography list.
% -------------------------------------------------------------------------
\bibliographystyle{IEEEbib}
\bibliography{conf}

\end{document}